# Vision-Guided Closed-Loop Autonomous Navigation of Light-Driven Janus Microbots

Kanhu Charan Andia[1] | Harsh Prashant Katole[2] | Parth Nitinkumar Bhosale[3] | Hemant Kumar[1]| Dhruv Pratap Singh[1*]

[1] *Department of Physics, IIT Bhilai, Durg, Chhattisgarh, 491002, India*
[2] *Department of Mechatronics, IIT Bhilai, Durg, Chhattisgarh, 491002, India*
[3] *Department of Computer science and Engineering, IIT Bhilai, Durg, Chhattisgarh, 491002, India*

[*] Corresponding Author: dhruv@iitbhilai.ac.in

## Abstract

Artificial microbots capable of precise and autonomous navigation could provide versatile platforms for microscale transport, manipulation, and task execution in complex environments. However, reliable navigation becomes increasingly challenging at the micrometre scale, particularly when the surrounding environment changes dynamically. Here, we develop a vision-guided closed-loop navigation framework for light-driven cube-shaped hematite/$TiO_2$ Janus microbots with a characteristic size of approximately 3 μm. Real-time visual localization using a custom-trained YOLOv11 object-detection model is integrated with discretized waypoint-based magnetic steering to enable precise navigation along complex prescribed trajectories, with more than 90% of the measured trajectory deviations remaining below 6 μm across the tested path geometries. The framework further enables arbitrary user-defined free-hand trajectories and autonomous navigation through a physically confined maze. In dynamically changing environments, A* path planning enables real-time path replanning while avoiding multiple independently swimming 5 μm spherical Janus micromotors. Finally, the microbot selectively captures multiple cargo particles and autonomously transports and delivers them to a predefined location while simultaneously avoiding dynamic obstacles. These results establish an integrated platform combining programmable trajectory control, autonomous path planning, real-time environmental adaptation, and task execution within a single light-driven Janus microbot system.

**Supporting information available:** Videos S1–S8 and Figures S1–S18.

## 1 | Introduction

Microbots capable of controlled motion in fluidic environments have attracted considerable attention for applications in targeted delivery, microscale manipulation, sensing, environmental remediation, and lab-on-a-chip technologies[1–6]. Their ability to perform such functions depends not only on propulsion but also on precise control over their position and direction of motion[7–9]. A variety of propulsion and actuation mechanisms based on chemical reactions[10,11], light[12–15], magnetic fields[16–20], acoustic waves[21–23], and electric fields[24,25] have therefore been developed, enabling microbots with diverse modes of motion and functionality[26–33]. However, reliable navigation becomes increasingly challenging at the micrometre scale, where Brownian fluctuations, local interactions, and variations in the surrounding environment can strongly influence microbot motion[34–38]. Developing approaches that can precisely control microbot trajectories while retaining sufficient flexibility to perform different navigation tasks is therefore an important requirement for their practical operation.

Self-propulsion alone, however, does not provide reliable control over where a microbot moves or how accurately it follows a desired path. Closed-loop control addresses this limitation by continuously determining the microbot position and applying corrective actuation according to the prescribed trajectory[39–42]. Vision-based feedback is particularly attractive for this purpose because it enables real-time localization without requiring additional onboard components. Such feedback can be combined with programmable trajectory generation to guide microbots along mathematically defined paths or arbitrary user-defined trajectories[43,44]. An effective navigation framework should therefore provide not only positional correction but also sufficient programmability to translate different trajectory inputs into reproducible microscale motion. This becomes particularly important for complex trajectories containing frequent changes in direction, where small deviations can accumulate and substantially affect path-following accuracy. Beyond predefined trajectories, intuitive generation of arbitrary paths can further increase the flexibility of microbot navigation by allowing desired motion to be specified directly by the user without requiring individual mathematical descriptions for each trajectory.

Recent studies have progressively expanded closed-loop microbot navigation beyond simple trajectory following. Li et al. demonstrated autonomous collision-free navigation of chemically propelled Janus microvehicles using vision-based localization and adaptive path planning in complex and dynamically changing environments[45]. Kirmizitas et al. subsequently demonstrated closed-loop control of acoustically propelled, magnetically steerable microrobots along predefined trajectories, together with precise positional control and cell manipulation[46]. Gauri et al. integrated real-time sensing, path planning, and magnetic actuation to achieve autonomous navigation of microellipsoids through maze environments together with cargo capture and transport[47]. More recently, Rachbuch et al. demonstrated hybrid magnetic–electric actuation for automated closed-loop navigation, multi-surface operation, and cargo transport of Janus microrobots[48], while learning-based approaches, including the work of An et al., have extended autonomous navigation to microrobotic swarms operating in unknown and dynamically changing environments[49]. These studies highlight substantial progress toward autonomous microscale operation. Nevertheless, integrating precise and programmable trajectory control with environment-aware path generation and real-time adaptation within an individual light-driven microbot at the micrometre scale remains challenging. Of particular interest is a framework in which the same microbot can progress from accurately executing programmed or directly user-defined trajectories to navigating through physical confinement and subsequently reconfiguring its path in response to independently moving microscale objects. Performing a physical task while maintaining such adaptive navigation places additional demands on the integration of visual perception, path planning, and actuation.

Here, we establish a vision-guided closed-loop autonomous navigation framework, termed VIMAN (Vision-guided Microbot Autonomous Navigation), for light-driven hematite/titania Janus microbots with a characteristic size of approximately 3 μm. The cube-shaped microbots consist of hematite asymmetrically coated with $TiO_2$, enabling light-driven propulsion in an aqueous hydrogen peroxide medium, while the magnetic response of hematite provides directional steering using a two-dimensional magnetic field. Real-time microscopic images are analyzed using a custom-trained YOLOv11 object-detection model together with a custom-developed Python control program to determine the positions of the microbot and surrounding obstacles. Prescribed trajectories are discretized into equidistant waypoints and executed through closed-loop magnetic steering, while integration of a graphics-tablet-based interface enables arbitrary free-hand paths to be directly translated into executable microbot trajectories. We further extend the framework to autonomous navigation through a physically confined maze using image-based boundary detection and to dynamically changing environments, where A* path planning enables real-time replanning around independently moving obstacles. Finally, the framework is applied to selective and sequential cargo capture, transport, and delivery to a predefined location while simultaneously avoiding dynamic obstacles. The overall VIMAN concept and its integration of programmable trajectory execution, environment-aware path planning, and task-oriented operation are schematically illustrated in **Figure 1**.

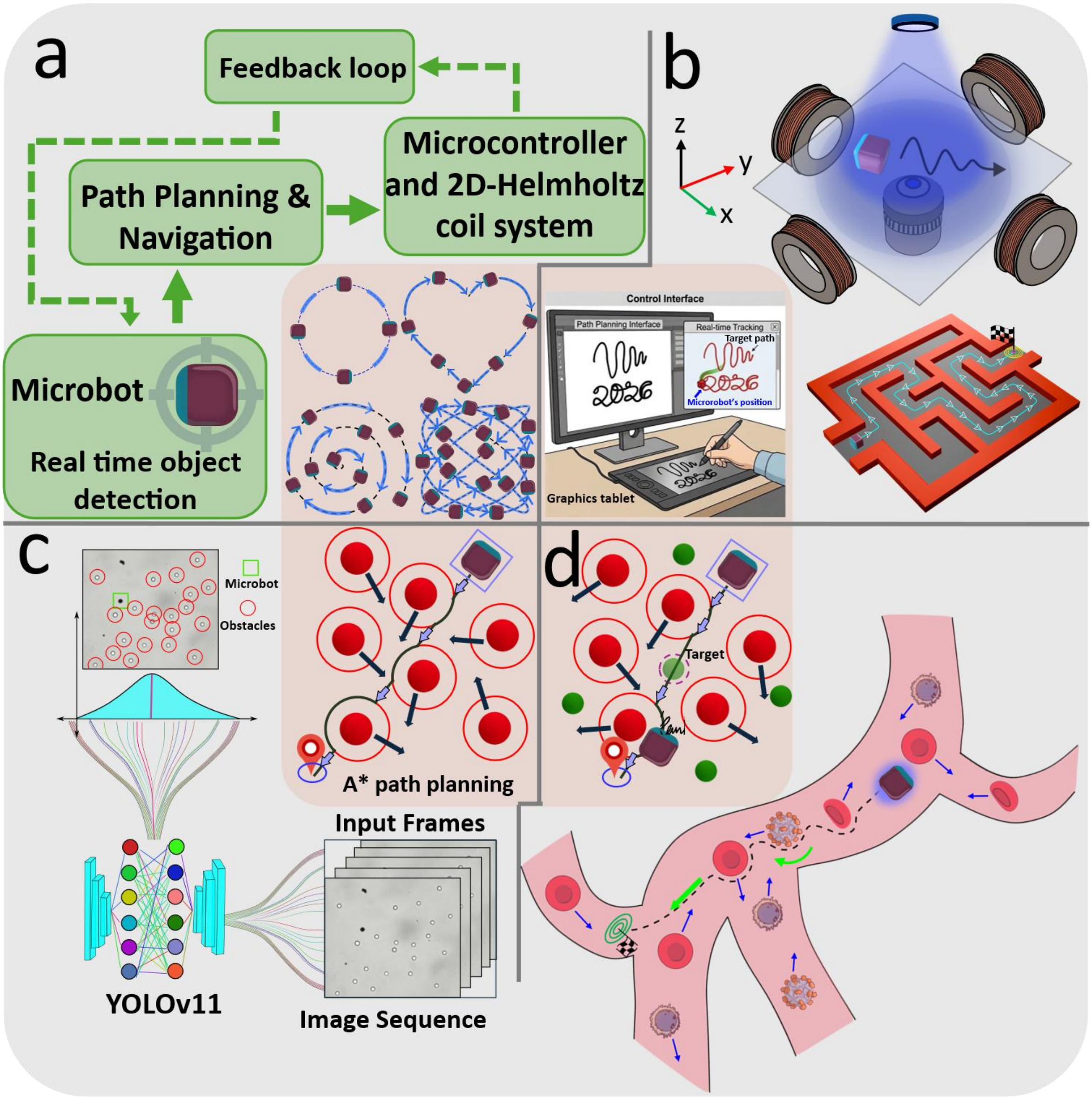

**Figure 1.** Overview of the VIMAN framework for autonomous navigation of light-driven Janus microbots. (a) Closed-loop workflow integrating real-time object detection, path planning, magnetic steering, and feedback control. (b) Experimental platform and navigation in physically confined environments. (c) YOLOv11-based object detection and A* path planning for dynamic obstacle avoidance. (d) Autonomous cargo capture, transport, and delivery in complex environments.

## 2 | Results and discussion

### 2.1 | Light-Driven Hematite/Titania Janus Microbots and Magnetic Steering

We first developed cube-shaped hematite/titania Janus microbots with an edge length of approximately 3 μm as the active platform for autonomous microrobotic navigation (**Figure 2a**). The microbots consist of a hematite (α-$Fe_2O_3$) cube asymmetrically coated with a titania ($TiO_2$) layer on one exposed face, forming a Janus architecture. The inset scanning electron microscopy (SEM) image in **Figure 2a** shows the cubic morphology of the hematite/titania microbots. The fabrication procedure is described in the *Experimental Section* and in **Figure S1**, Supporting Information. Additional structural, optical, and material characterization, including EDX, X-ray diffraction, UV–Vis spectroscopy, and vibrating sample magnetometry (VSM), are provided in **Note 2.2**, Supporting Information. This Janus architecture provides the microbots with distinct photo-responsive and magnetic properties, which can be exploited for propulsion and directional control, respectively.

Under blue-light illumination (465 nm), the hematite/titania Janus microbots exhibited spontaneous propulsion in a low-concentration aqueous hydrogen peroxide medium (0.5% w/v) (**Figure 2b**). In the absence of illumination, the microbots displayed only Brownian motion, confirming the light-dependent nature of the propulsion. Under illumination, the microbots exhibited an average swimming speed of approximately 5-6 μm $s^{-1}$ and moved spontaneously with continuously changing directions. As schematically illustrated in **Figure 2b**, blue-light irradiation generates photogenerated charge carriers in the hematite/titania system, which participate in photocatalytic reactions involving $H_2O_2$. Interfacial charge-transfer processes and the asymmetric Janus structure produce spatially nonuniform reaction rates around the microbot, resulting in local concentration gradients and phoretic slip that drive diffusiophoretic propulsion[50–52]. The dependence of the swimming speed on $H_2O_2$ concentration and incident light intensity is shown in the **Figure S3a,b,** Supporting Information, demonstrating the tunability of the propulsion under the experimental conditions used here.

To regulate this otherwise unconstrained or random motion, we exploited the magnetic responsiveness of the hematite component using a two-dimensional Helmholtz-coil system (**Figure 2c**). Two orthogonal pairs of coils generate magnetic fields along the x- and y-directions, enabling control of the microbot's orientation within the imaging plane. By varying the applied magnetic field, the direction of the light-driven microbot could therefore be controlled while maintaining optical propulsion[53,54]. This separation of propulsion and directional control, in which blue light drives the microbot while magnetic actuation determines its direction of motion, provides the physical basis for the closed-loop navigation framework developed in the following sections.

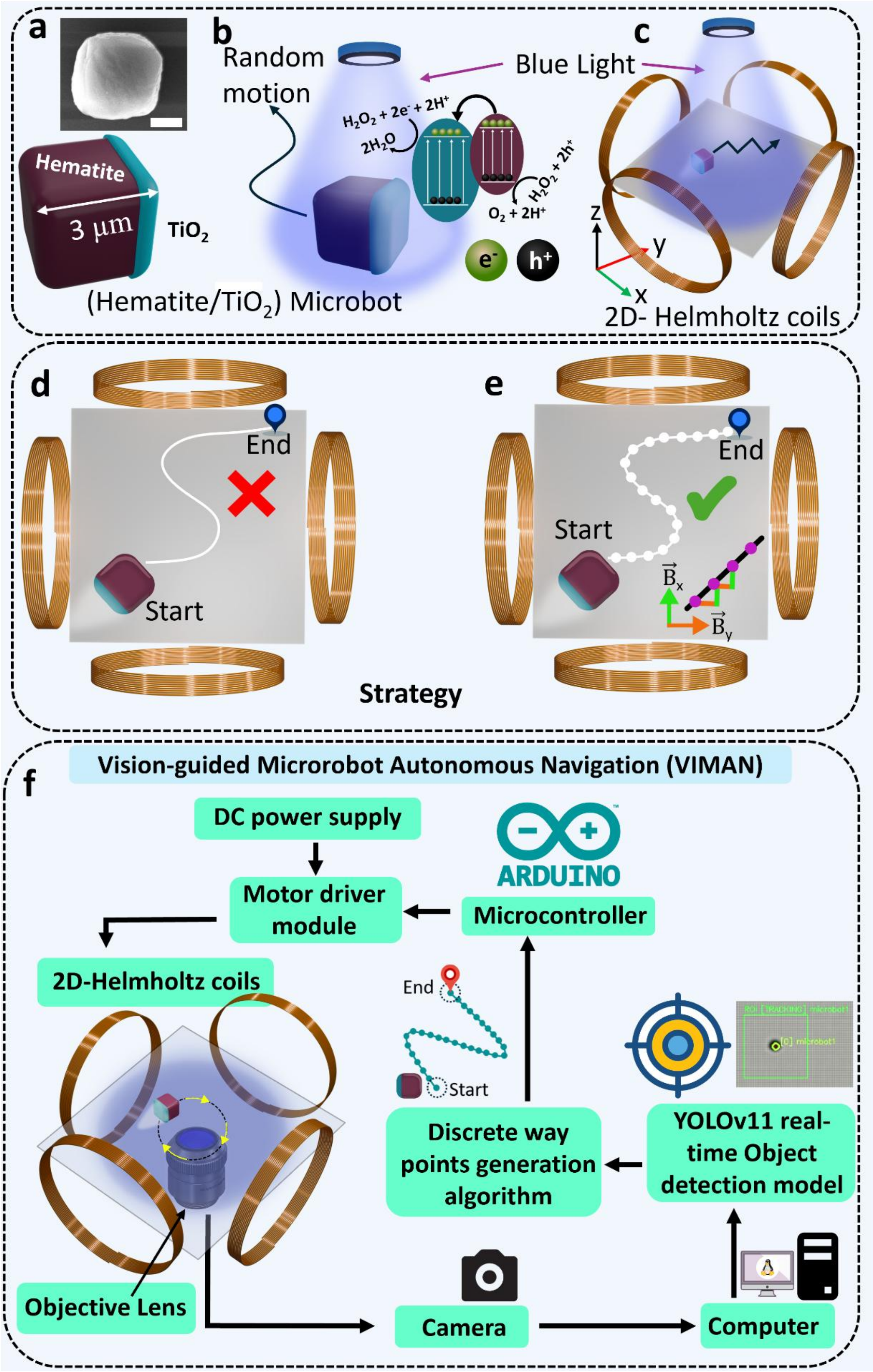

**Figure 2.** Light-driven Janus microbot and VIMAN closed-loop navigation framework. (a) Schematic and SEM image of the hematite/$TiO_2$ Janus microbot. (b) Schematic of light-driven random motion and photocatalytic propulsion under blue-light illumination in 0.5% (w/v) aqueous $H_2O_2$. (c) 2D Helmholtz-coil setup for magnetic steering. (d,e) Comparison of direct trajectory following and waypoint-based discretization for closed-loop navigation. (f) VIMAN framework for vision-guided closed-loop control along programmable trajectories.

## 2.2 | Vision-Based Closed-Loop Navigation via Discrete Waypoint Control

Although magnetic actuation enables directional steering of the light-driven microbots, accurate navigation along complex trajectories requires continuous knowledge of the microbot position and real-time correction of its motion. We therefore developed an external vision-based closed-loop control system in which the microbot, cargo particles, and obstacles are identified using a custom-trained YOLOv11 object-detection model. The detection model exhibited high accuracy across the relevant particle classes (see **Note 2.4**, Supporting Information), enabling reliable extraction of object positions for real-time feedback control. The detected positions are continuously processed by the computer-based control program to determine the required steering commands, which are subsequently transmitted to the microcontroller for magnetic actuation. This provides continuous positional feedback and correction during navigation. Further details of the feedback-control strategy are provided in **Note 2.5**, Supporting Information.

To convert a desired trajectory into an experimentally executable path, we developed a path-discretization strategy in which the prescribed continuous trajectory (**Figure 2d**) is divided into a sequence of equidistant waypoints (**Figure 2e**). Rather than attempting to reproduce the complete trajectory continuously, the microbot is guided sequentially toward individual waypoints. The uniformly spaced waypoints provide repeated opportunities for positional correction throughout the motion, thereby limiting deviations from the prescribed trajectory.

For navigation between successive waypoints, we implemented a base-to-perpendicular steering strategy (see inset, in **Figure 2e**). The instantaneous microbot position is treated as the base point, while the next waypoint defines the target direction. Based on their relative positions, the required magnetic steering direction is determined, and the corresponding x- and y- direction Helmholtz coil actuation commands are generated. This geometric approach provides a computationally simple method for updating the steering direction during real-time navigation. Once the microbot reaches the prescribed waypoint within a defined positional tolerance, the controller advances to the next waypoint and repeats the process until the destination is reached.

The complete implementation of the VIMAN framework is illustrated in **Figure 2f** (see also **Note 2.6**, Supporting Information). Images acquired using the microscope camera are processed by the YOLOv11-based vision model to determine the instantaneous microbot position, while the control program determines the appropriate steering command based on its instantaneous position and the next target waypoint. The resulting command is transmitted through the Arduino-based control system to the motor-driver module and orthogonal Helmholtz coils, generating the magnetic field required to steer the microbot. The microbot position is then updated from the subsequent acquired image and returned to the control program as feedback, thereby closing the perception–control–actuation loop.

Desired trajectories were generated using a custom Python-based program that accepts either user-defined points or mathematical equations describing different geometries**.** The generated trajectories were discretized into equidistant waypoints and executed using the same closed-loop steering strategy. This programmable architecture allows trajectories with different geometries to be implemented without modifying the underlying control algorithm, providing the basis for the trajectory-following experiments described in the following section.

## 2.3 | Precise Navigation along Complex Prescribed Trajectories

The developed closed-loop system was first evaluated for its ability to accurately reproduce prescribed trajectories spanning a broad range of geometries. The desired trajectories were generated using a custom Python-based program that accepts user-defined points as well as mathematical equations for different geometries (see **Note 2.7**, Supporting Information). Using the same control architecture, the microbot successfully navigated along a diverse range of prescribed trajectories, including polygonal, circular, lemniscate (∞), spiral, heart-shaped, and Lissajous paths (**Figures 3a–f**). In each case, the prescribed trajectory was discretized into equidistant waypoints, and the microbot was sequentially guided through these waypoints using closed-loop magnetic steering. The green traces represent the prescribed waypoint-based trajectories, whereas the corresponding right-hand panels show the experimentally recorded trajectories, with the microbot (black square) visible along the tracked path. A representative time-lapse sequence of the microbot following a prescribed trajectory is shown in **Figure 3g**. The close correspondence between the prescribed and experimentally followed trajectories demonstrates the ability of the proposed approach to reproduce substantially different path geometries within the same control framework.

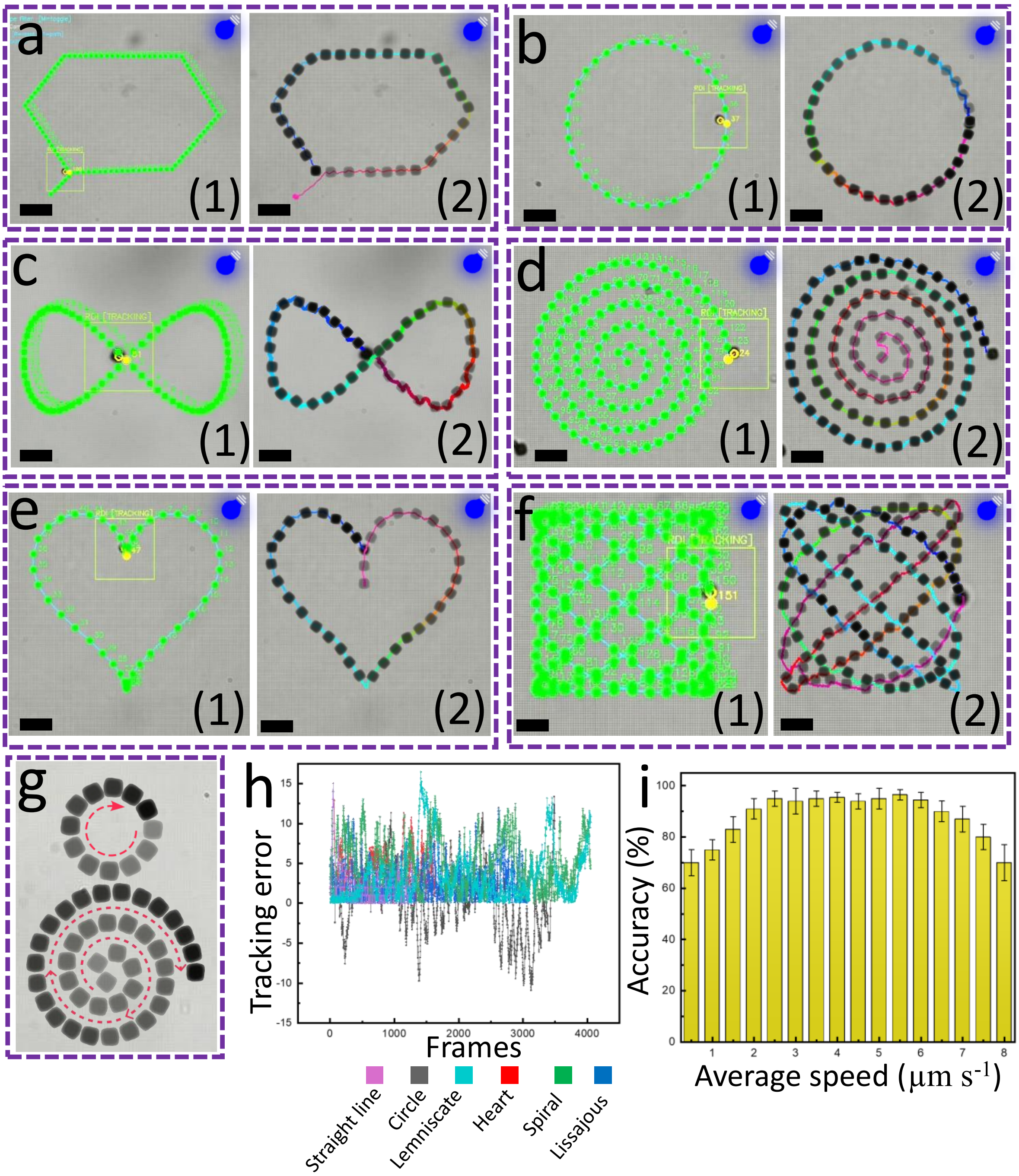

**Figure 3.** Closed-loop navigation along complex prescribed trajectories. (a–f) Programmed and experimentally recorded trajectories for (a) polygonal, (b) circular, (c) lemniscate, (d) spiral, (e) heart-shaped, and (f) Lissajous paths under blue-light illumination in 0.5% (w/v) aqueous $H_2O_2$. For each trajectory, (1) shows the programmed path and (2) the corresponding experimentally followed microbot trajectory. (g) Representative schematic/time-lapse illustration of microbot motion along circular and spiral paths, with faded and solid microbot positions indicating the start and end positions, respectively. (h) Trajectory-tracking error as a function of frame/time. (i) Trajectory-following accuracy as a function of average swimming speed. Scale bars, 10 μm.

The experimentally recorded trajectories remained closely aligned with their corresponding prescribed paths across the tested geometries, despite the continuously changing steering directions required for increasingly complex trajectories. In particular, the lemniscate, spiral, heart-shaped, and Lissajous trajectories involved repeated changes in steering direction, and curvature, while the Lissajous trajectory (frequency ratio of 5:4) additionally contained closely spaced path segments**.** Nevertheless, the microbot successfully traversed these trajectories using the same closed-loop control architecture. This ability to maintain correspondence with the prescribed paths during complex directional changes demonstrates that the discretized waypoint-based feedback strategy can continuously correct positional deviations during navigation without requiring a geometry-specific control strategy.

To quantitatively assess trajectory-following accuracy, the deviation between the prescribed and experimentally recorded trajectories was evaluated using trajectory-error maps, path-deviation analysis, error distributions, polar error maps, cumulative root-mean-square error (RMSE), and box plots (see **Note 2.8**, Supporting Information). The cumulative RMSE converged to approximately 2-3 μm for the linear reference trajectory and approximately 5–6 μm for the more complex trajectories shown in **Figure 3a–f**. Importantly, more than 90% of the measured trajectory deviations remained below 5 μm across the tested path geometries, demonstrating that the microbot remained closely confined to the prescribed paths despite repeated changes in steering direction.(**Figure 3h**). Although the magnitude and distribution of the tracking error varied among the different trajectories, the deviations remained bounded throughout navigation. We further examined the dependence of trajectory-following accuracy on the swimming speed of the microbot (**Figure 3i**). The microbot maintained consistent trajectory-following accuracy across the tested speed range, indicating that the waypoint-based feedback strategy remains effective over different operating speeds, (see **video S1**, Supporting Information). Overall, these results demonstrate that the proposed control strategy provides reliable trajectory tracking across substantially different path geometries and operating speeds.

### 2.4 | Programmable Navigation along User-Defined Free-Hand Trajectories

Having established precise navigation along predefined geometric trajectories, we next investigated whether the VIMAN navigation framework could translate arbitrary user-drawn free-hand paths into executable microbot trajectories. For this purpose, a graphics-tablet-based free-hand path-planning interface was integrated with the VIMAN framework (**Figure 4a**). The user can draw a desired trajectory directly on the graphics tablet, after which the recorded path is processed in real time to generate a sequence of discretized equidistant waypoints within the microscopic image interface. These waypoints are subsequently supplied to the closed-loop navigation system, where the microbot follows the prescribed free-hand trajectory through sequential magnetic steering.

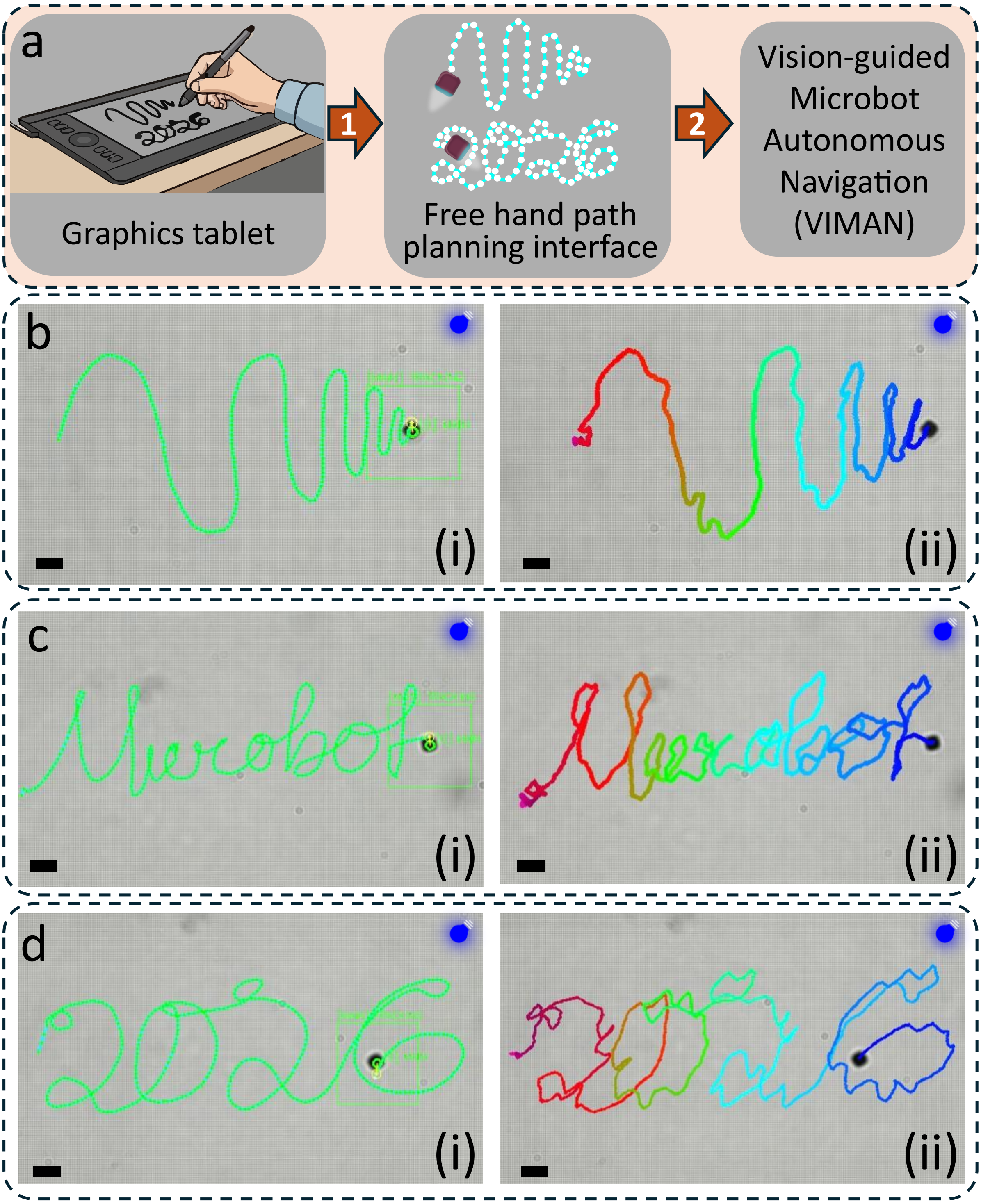

**Figure 4.** Programmable navigation along user-defined free-hand trajectories. (a) Graphics-tablet interface for drawing free-hand paths, generating discretized waypoints, and transferring them to the VIMAN framework. (b–d) User-defined trajectories corresponding to (b) a damped sine-wave pattern, (c) the handwritten word "Microbot," and (d) the handwritten number "2026." For each case, (i) shows the user-drawn trajectory and (ii) the corresponding experimentally followed microbot trajectory under blue-light illumination. Scale bars, 10 µm.

Several free-hand trajectories with substantially different shapes were used to evaluate this capability (**Figures 4b–d**). The user-drawn trajectories are shown in the left panels (**Figures 4b(i)–d(i)**), while the corresponding experimentally recorded microbot trajectories are shown in the right panels (**Figures 4b(ii)–d(ii)**). The microbot successfully reproduced different free-hand trajectories, including a damped sine-wave pattern, the handwritten word "Microbot," and the handwritten number "2026", together with additional shapes shown in **Videos S2** and **S3** (Supporting Information). Importantly, the same trajectory-generation and closed-loop control architecture was used for all inputs, without requiring geometry-specific mathematical equations or modification of the underlying navigation strategy.

These results demonstrate that the VIMAN framework can directly translate user-drawn free-hand paths into executable microscale trajectories, providing an intuitive approach for programmable and task-specific microbot navigation without requiring the user to define the underlying mathematical trajectory or steering commands.

### 2.5 | Autonomous Navigation and Maze Solving in Physically Confined Environments

Following the successful execution of complex trajectories in an unobstructed workspace, we next investigated the ability of the microbot to navigate through physically confined environments. A maze structure with predefined boundaries was used to provide a controlled environment for evaluating autonomous navigation under spatial constraints. Unlike the previously demonstrated trajectory-following experiments, where the desired path was prescribed in advance, the navigation path in the maze had to be generated according to the detected environmental geometry while avoiding the surrounding walls (**Figure 5a**).

The maze was fabricated using the standard positive photolithography technique (see **Figure S12**, Supporting Information). Its boundaries were identified from the acquired microscopic images using a threshold-based detection approach, in which the black straight-line regions were segmented and treated as maze walls (see **Note 2.9,** Supporting Information). The maze consisted of interconnected passages with a characteristic width of approximately 20 μm, providing a confined environment with multiple turns and restricted passages. As shown in **Figure 5b**, the initial microbot position and the designated destination were first defined within the detected maze geometry. Based on this information, the custom-developed navigation algorithm generated an executable collision-free path connecting the microbot to the destination while avoiding the maze walls (**Figure 5c**). The generated path was subsequently discretized into waypoints and supplied to the closed-loop magnetic steering system. The microbot successfully followed the generated path through the confined passages and successive changes in direction (**Figures 5d,5e)** and reached the designated destination without colliding with the surrounding maze walls (**Video S4**, Supporting Information).

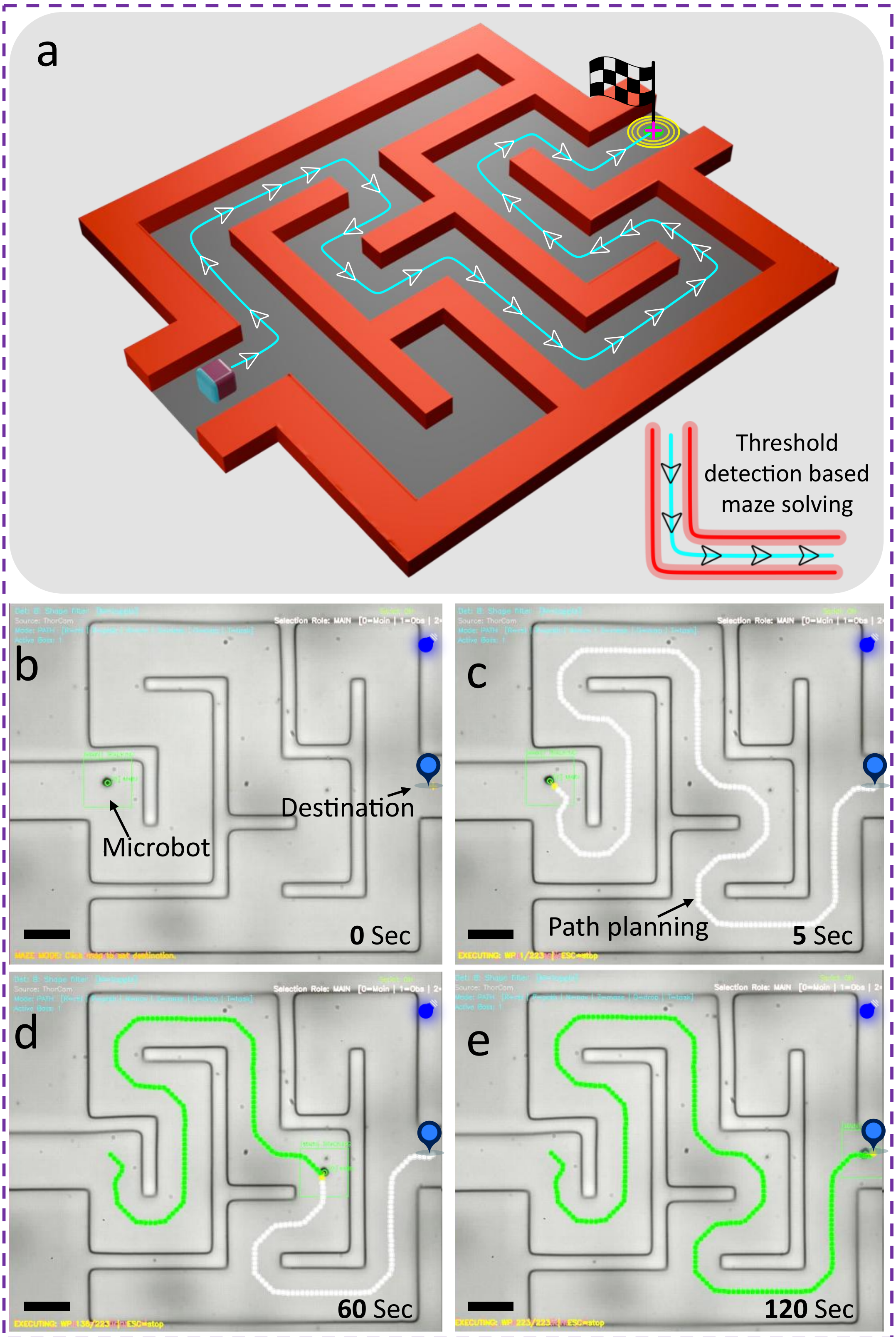

**Figure 5.** Autonomous maze navigation in a physically confined environment. (a) Schematic of maze solving using real-time threshold-based detection of the maze boundaries. (b) Microscopic image showing real-time detection of the microbot and selection of the destination within the maze. (c) Threshold-detection-based path generation through the maze. (d) Intermediate stage of navigation, with green dots indicating the path already traversed by the microbot. (e) Successful completion of the planned path and arrival of the microbot at the predefined destination. Scale bars, 15 μm.

This experiment demonstrates that the VIMAN framework can extend beyond reproduction of prescribed trajectories to environment-aware path generation and autonomous navigation within physically confined surroundings. The successful traversal of the maze through multiple turns and restricted passages demonstrates the ability of the system to incorporate the detected environmental geometry into navigation, providing a foundation for autonomous navigation in more complex environments containing dynamically changing surroundings.

### 2.6 | Real-Time Adaptive Navigation in Dynamic Environments

Having demonstrated accurate trajectory following in an unobstructed workspace and navigation within physically confined environments, we next investigated whether the microbot could adapt its trajectory in response to dynamically changing surroundings. For the main dynamic obstacle-avoidance experiments, approximately 5 μm spherical $Cu_2O/SiO_2$ Janus micromotors with an average swimming speed of 4-5 μm $s^{-1}$ were used as independently swimming active obstacles. The overall navigation strategy is schematically illustrated in **Figure 6a**. The microbot was required to reach a predefined destination while continuously adapting its path to the changing positions of multiple moving obstacles using reconfigurable A* path planning [55]. Details of the A* path-planning strategy are provided in **Note 2.10.1**, Supporting Information.

Representative microscopic images of the active dynamic obstacle-avoidance process are shown in **Figure 6b–e**. At the beginning of navigation, the microbot and dynamic obstacles were identified from the acquired microscopic image using the custom-trained YOLOv11 object-detection model, while the destination was predefined. An initial path toward the destination was then generated using A* path planning while accounting for the detected obstacle positions. A circular safety zone with a radial extent of approximately 10 μm was defined around each detected obstacle and incorporated into the path-planning process. As the obstacles continuously changed their positions, the system repeatedly evaluated the safety of the current path. When the updated obstacle positions rendered the current path unsafe, a new collision-free A* path was generated in real time while accounting for the updated obstacle safety zones (see **Note 2.10.2**, Supporting Information) in real time. As shown in **Figure 6b,c**, the initially generated path was modified in response to the changing obstacle positions, after which the microbot followed the newly generated path while continuing toward the same destination (**Figure 6d**). Multiple such path adaptations occurred during navigation as different obstacles approached the microbot's planned route. The rapid reconfiguration of the navigation path demonstrates the ability of the closed-loop system to respond to changes in the surrounding environment in real time (see **Note 2.10.3**, Supporting Information).

Throughout these experiments, no collision between the microbot and the moving obstacles was observed. The successful avoidance of multiple independently moving obstacles demonstrates that the proposed framework can integrate real-time object detection, obstacle safety constraints, dynamic A* path replanning, and closed-loop magnetic steering within a changing environment. Rather than following a fixed prescribed trajectory, the microbot continuously adapts its navigation path according to the updated obstacle positions while maintaining the same destination. As shown in **Figure 6e**, the microbot successfully reached the designated destination after navigating through the dynamically changing obstacle field (see video S5, Supporting Information).

To further evaluate navigation in dynamically changing environments, additional experiments were performed using 5 μm $SiO_2$ particles undergoing Brownian motion as dynamic obstacles (**Figure S14 & 15; Video S6**, Supporting Information), providing a complementary test of navigation in fluctuating surroundings. Additional navigation experiments involving multiple predefined target locations are shown in **Video S7**, Supporting Information. The influence of the obstacle safety-zone size on collision avoidance performance was also evaluated, as shown in **Figure S16**, Supporting Information. Together, these experiments demonstrate adaptive and collision-free navigation under a range of dynamically changing microscale conditions.

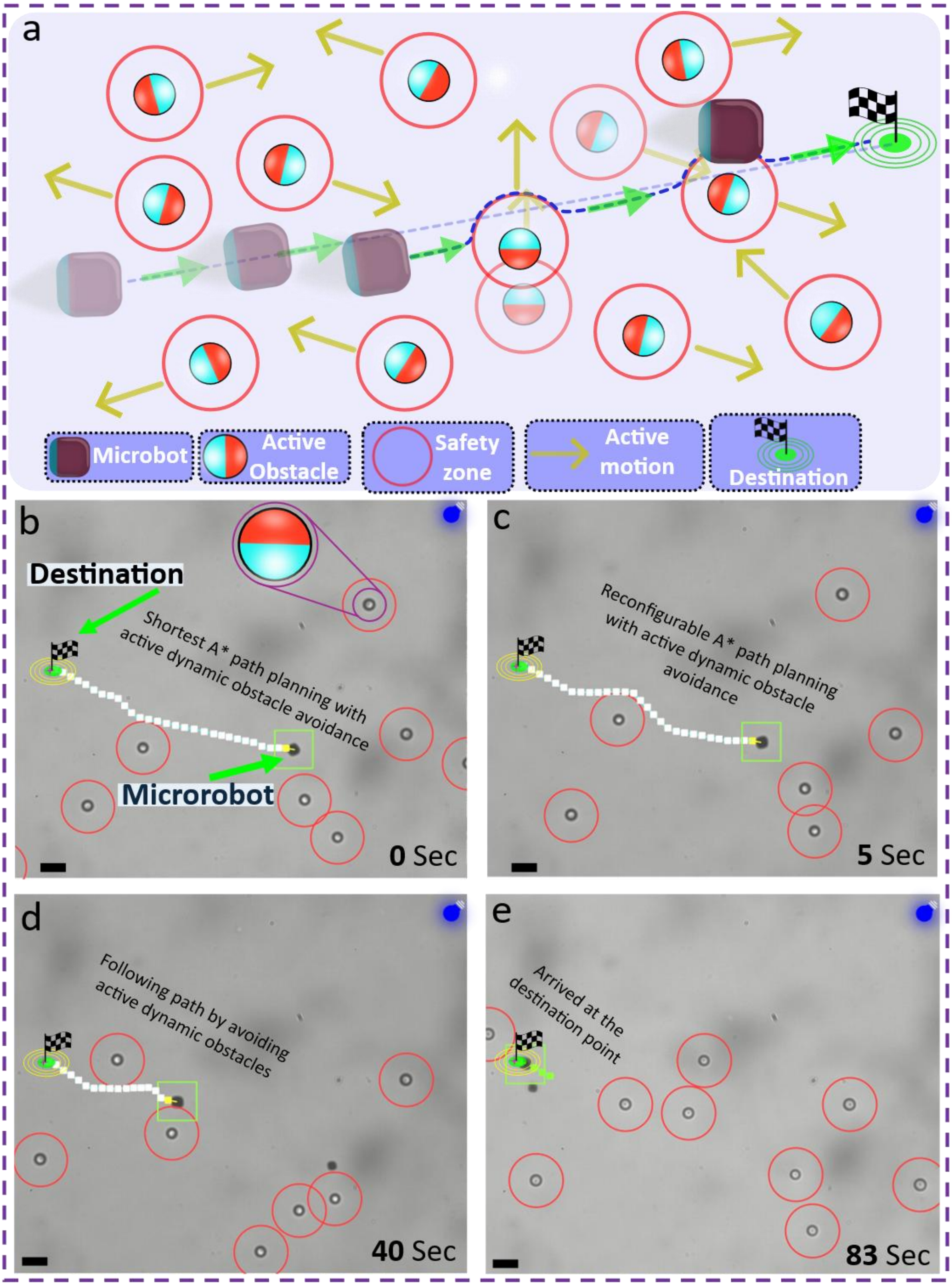

**Figure 6.** Real-time adaptive navigation in a dynamic environment. (a) Schematic of VIMAN-based avoidance of actively moving obstacles. (b–e) Representative microscopic images showing microbot detection and destination selection, initial A* path generation, real-time path replanning in response to moving obstacles, and successful arrival of the microbot at the predefined destination. Scale bars, 10 μm.

## 2.7 | Autonomous Cargo Transport through Dynamic Environments

Having established collision-free navigation in dynamically changing environments, we next investigated whether the microbot could perform a complete cargo-transport task involving target selection, sequential cargo capture, dynamic obstacle avoidance, transport, and delivery to a predefined location**.** As illustrated schematically in **Figure 7a**, multiple cargo particles were designated as targets, and the microbot was programmed to sequentially collect these particles and transport them to a predefined delivery location while navigating through a dynamically changing environment.

The custom-trained YOLOv11 object-detection model simultaneously identified the microbot, 5 μm $SiO_2$ particles used as dynamic obstacles, and 2 μm PLA particles used as cargo, while the delivery location was predefined. The positions of the microbot, cargo particles, and obstacles were continuously updated during navigation as the cargo and obstacle particles underwent Brownian motion. Following object detection, the A* path-planning algorithm generated a collision-free path toward the first selected target while accounting for the surrounding obstacles. The microbot then approached the target along the generated path. The local chemical gradient generated around the microbot induces diffusiophoretic interactions with the cargo particle, enabling capture when the microbot comes into close proximity to the target[56,57]. Once the selected cargo particle was captured, it was marked as collected and the existing path was cleared, allowing a new obstacle-aware path to be generated toward the next selected target **(**see **Note 2.11.1** & **2.11.2**, Supporting Information). This process was repeated until the prescribed cargo particles had been collected, after which a final path was generated toward the predefined delivery location.

Representative microscopic images of this process are shown in **Figure 7b–g.** At the beginning of the experiment, the microbot, target cargo particles, and surrounding obstacles were identified, and an obstacle-aware path toward the first selected target was generated (**Figure 7b,c**). The microbot subsequently approached and captured the first target before being redirected toward the second target (**Figures 7d,e**). After collecting the second target, the microbot transported the captured cargo toward the predefined delivery location while continuously accounting for the changing obstacle positions (**Figure 7f**). Upon reaching the predefined delivery location, rapid programmed switching of the external magnetic field induced repeated reorientation of the microbot. This reorientation continuously altered the relative position of the active surface with respect to the captured cargo particles, destabilizing the local phoretic interaction responsible for cargo retention and thereby enabling their on-demand release at the desired location. (**Figure 7g**, see **Note 2.11.3** & **video S8,** Supporting Information). For the cargo-transport experiments, UV illumination was used to enhance the phoretic interaction between the microbot and the cargo particles; further details are provided in **Note 2.12**, Supporting Information. Although the representative experiments shown here were performed using 2 μm PLA particles as model cargo, additional experiments confirmed capture and transport of 2 μm silica particles ($SiO_2$), demonstrating that cargo-transport capability is not restricted to a single cargo material (see **Figure S18**, Supporting Information).

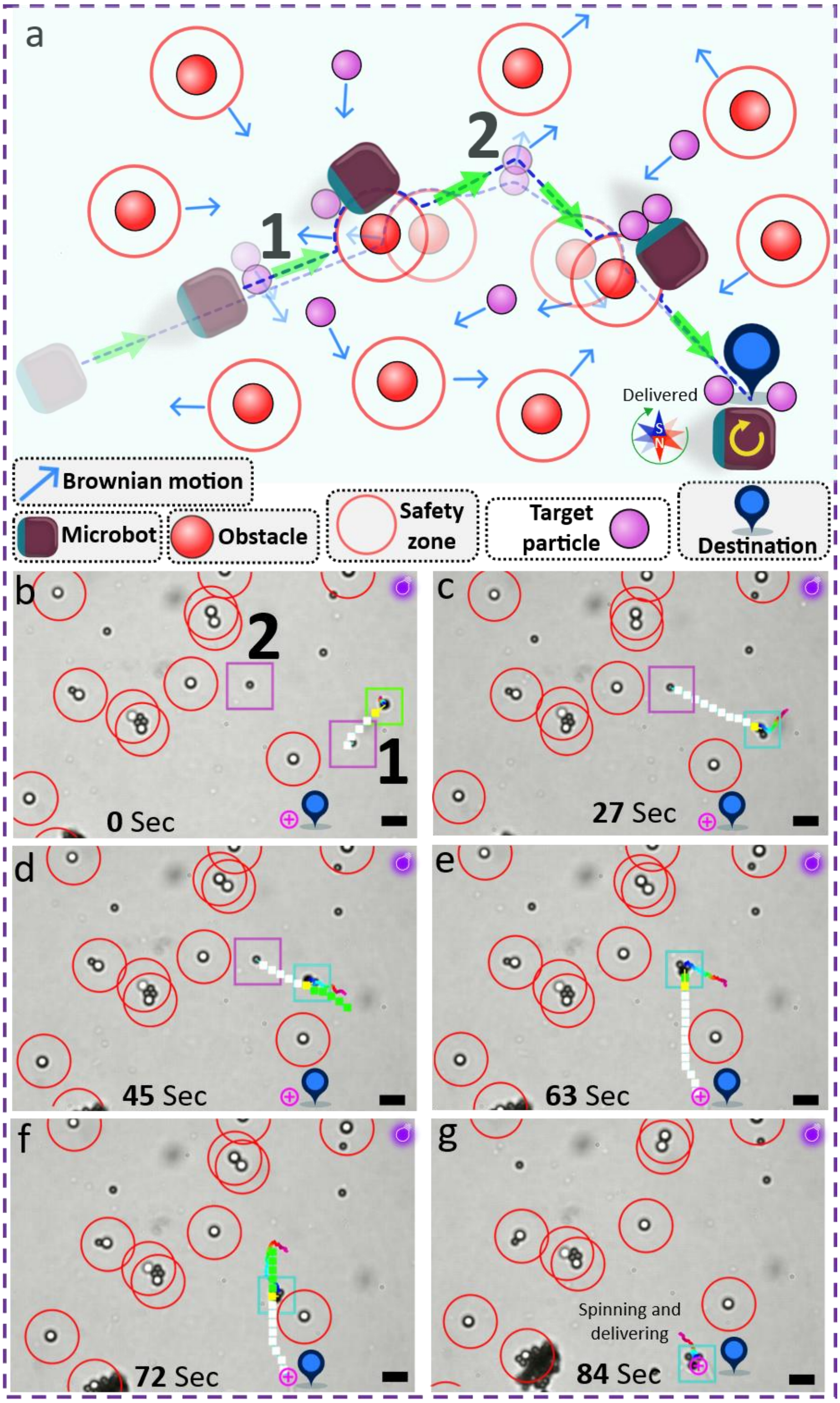

**Figure 7.** Autonomous multi-cargo transport in a dynamic environment. (a) Schematic of VIMAN-based sequential capture and transport of 2 µm PLA cargo particles, including real-time object detection, dynamic obstacle avoidance, A* path replanning, and magnetically triggered cargo release. (b–g) Representative microscopic images showing target and delivery-point selection, capture of the first and second cargo particles, transport toward the predefined delivery location while avoiding dynamic obstacles, and final cargo release by rapid external magnetic-field switching under UV illumination. Scale bars, 10 µm.

This experiment demonstrates the integration of real-time object detection, target selection, dynamic path planning, sequential cargo capture, obstacle avoidance, transport, and controlled release within a single autonomous navigation framework. The ability to selectively collect multiple dynamically positioned cargo particles and deliver them to a predefined location while continuously navigating around moving obstacles extends VIMAN framework from autonomous navigation alone to task-oriented microscale operation in dynamically changing environments.

## 3 | Experimental Section

### 3.1 | Fabrication of Hematite/Titania Janus Microbots

Hematite/Titania Janus microbots were fabricated using a two-step process involving hydrothermal synthesis of hematite microparticles followed by electron-beam evaporation of a titania ($TiO_2$) thin film. First, cube-shaped hematite microparticles with an edge length of approximately 3 μm were synthesized using a previously reported hydrothermal method[58], with slight modifications to obtain larger hematite cubes. Briefly, 0.44 g of $FeCl_3$ was dissolved in 15 mL of deionized (DI) water under magnetic stirring. Subsequently, 15 mL of 0.02 M HCl was added dropwise through a syringe pump at a constant flow rate of 250 μl $min^{-1}$. The resulting solution was then stirred for 30 min at 200 rpm and transferred to a 100 mL Teflon-lined stainless-steel autoclave. The hydrothermal reaction was carried out at 180 °C for 24 h, after which the autoclave was allowed to cool naturally to room temperature (approximately 25–30 °C). The resulting hematite microparticles were collected by centrifugation and washed 3–4 times with ethanol and DI water.

For the formation of the Janus structure, the synthesized hematite cubes were dispersed, drop-cast onto a silicon wafer, and dried on a hot plate at approximately 60 °C. The dried sample was subsequently transferred to the vacuum chamber of an electron-beam evaporation system, where the chamber pressure was maintained at approximately $10^{-6}$ mbar. A $TiO_2$ layer with a thickness of approximately 160 nm was deposited onto the exposed face of the hematite microparticles, producing the asymmetric hematite/titania Janus architecture[59,60]. The morphology and crystal structure of the synthesized hematite/titania Janus microbots were characterized using scanning electron microscopy (SEM) and X-ray diffraction (XRD), respectively. The optical absorption properties were investigated using UV–visible absorption spectroscopy. The magnetic properties of the hematite microparticles were evaluated using vibrating sample magnetometry (VSM), and the corresponding magnetization–magnetic field (M–H) hysteresis curves were obtained. The elemental composition and spatial distribution of Fe, Ti, and O were analyzed using energy-dispersive X-ray spectroscopy (EDX) coupled with SEM. Detailed characterization results are provided in **Figure S1**, Supporting Information.

### 3.2 | Sample preparation

Following $TiO_2$ deposition, the silicon wafer was sonicated in DI water for approximately 5–10 min to detach the hematite/titania Janus microbots from the wafer and disperse them in aqueous medium. Glass coverslips were plasma-cleaned for 2 min to render their surfaces hydrophilic. Subsequently, 4–6 µL of the microbot suspension was drop-cast onto the cleaned coverslip, followed by the addition of approximately 200 µL of 0.5% (w/v) aqueous hydrogen peroxide ($H_2O_2$) solution. The sample was then placed under an inverted optical microscope for observation and characterization of the microbot motion.

### 3.3 | Imaging and analysis

Imaging and visualization were performed using an inverted optical microscope (Leica DMi8) equipped with a Thorlabs scientific CMOS camera interfaced with a Linux-based computer through the ThorCam Python SDK. Images and videos were acquired in bright-field mode at a frame rate of 60 frames per second (fps). A white LED operated in transmission mode was used for bright-field illumination, while a blue laser was introduced through the epifluorescence illumination path to provide the optical excitation required for light-driven propulsion of the microbots in aqueous hydrogen peroxide ($H_2O_2$).

The recorded videos were analyzed using custom-developed Python scripts and open-source Python libraries. Trajectory visualization and quantitative analysis were performed using Origin together with custom Python-based analysis scripts**.** The same imaging framework was used for real-time detection, visualization, and tracking of the microbots, cargo particles, and obstacles, providing the image feedback required for the closed-loop navigation experiments.

### 3.4 | Maze Fabrication

The maze was fabricated on a square glass coverslip (24 × 24 $mm^2$) using a standard positive-maskless photolithography process. The coverslip was first plasma-cleaned for 5 min, followed by spin coating of a positive photoresist (S1813) at 3000 rpm for 30 s. The photoresist-coated coverslip was then pre-baked on a hot plate at 115 °C for 2 min. Subsequently, the substrate was transferred to a maskless photolithography system (MicroWriter ML3), where the desired maze pattern was defined by selective UV exposure**.** The exposed photoresist was developed using MF-319 developer, followed by post-baking at 120 °C for 3 min. The detailed fabrication procedure and maze design are provided in **Figure S12**, Supporting Information.

### 3.5 | Closed-loop Feedback control setup

Closed-loop guidance of the hematite/titania Janus microbots was achieved using a two-dimensional magnetic actuation system consisting of two pairs of Helmholtz coils positioned along the *x*- and *y*-axes of the inverted microscope. The coils were driven by a DC power supply (SCIENTIFIC PSD7303A) through a motor driver module (Cytron MDD20A), which was controlled using an Arduino Mega microcontroller.

The microcontroller was interfaced with a Linux based computer running a custom-developed Python control program. A custom YOLOv11 object-detection model, together with OpenCV-based image-processing routines, was integrated into the control framework for real-time identification and localization of the microbot, cargo particles, and obstacles within the workspace. The detected positional information was processed by a computer-based control program to determine the required magnetic actuation along the *x*- and *y*-directions. The corresponding commands were transmitted to the microcontroller, which controlled the motor-driver module and the magnetic fields generated by the Helmholtz coils. This closed-loop architecture established continuous feedback between visual detection and magnetic steering, enabling real-time correction of the microbot position and trajectory during navigation. A photograph of the experimental setup, including the Arduino-based control module, is shown in **Figure S9**, Supporting Information.

### 3.6 | Path Planning and Dynamic Obstacle Avoidance

The workspace was discretized into a grid based on the detected positions of the microbot, target, and obstacles, and the A* path-planning algorithm was used to generate a collision-free path between the microbot and the target. The generated path was subsequently converted into a sequence of waypoints for closed-loop magnetic steering. For dynamic obstacle avoidance, the positions of the moving obstacles were continuously updated from the acquired microscopic images, and the safety of the current path was evaluated in real time. When the updated obstacle positions rendered the current path unsafe, the A* algorithm was reinitiated to generate a new collision-free shortest path based on the updated obstacle configuration. To maintain continuity during replanning, the path-update procedure was implemented such that the immediate active waypoint of the microbot was not abruptly changed. The newly generated path was therefore connected to the ongoing waypoint sequence, preventing sudden changes in the steering command and allowing the microbot to transition smoothly to the replanned path. The updated path was then converted into waypoints and supplied to the closed-loop steering system, enabling real-time adaptation to changes in the surrounding environment.

### 3.7 | Obstacle Avoidance and Cargo Manipulation

For the dynamic obstacle avoidance experiment, approximately 5 μm $Cu_2O/SiO_2$ Janus micromotors were used as actively moving obstacles, while 5 μm $SiO_2$ particles were used for complementary experiments involving Brownian obstacles. The obstacle particles were first introduced onto a plasma-cleaned coverslip and dispersed in 0.5% w/v aqueous $H_2O_2$. After allowing the particles to sediment toward the coverslip, the hematite/titania Janus microbots were added to the same fluid medium. The positions of the microbot and obstacles were continuously detected during navigation using the custom-trained YOLOv11 object-detection model.

For the cargo-transport experiments, 2 μm diameter PLA and silica ($SiO_2$) microparticles were used as model cargo. The cargo particles were first introduced onto a plasma-cleaned coverslip and dispersed in 0.5% w/v aqueous $H_2O_2$. After sedimentation of the cargo particles, the hematite/titania Janus microbots were added to the fluid medium. Selected cargo particles were approached and captured by the microbot through local phoretic interactions and subsequently transported under closed-loop magnetic guidance. UV illumination was used during the cargo-transport experiments to enhance the phoretic interaction between the microbot and cargo particles. Cargo release at the predefined delivery location was achieved by rapid programmed switching of the external magnetic field, inducing repeated reorientation of the microbot and destabilizing the phoretic interaction responsible for cargo retention.

## 4 | Conclusion

We developed cube-shaped hematite/titania Janus microbots with a characteristic size of approximately 3 μm that exhibit light-driven propulsion under blue-light illumination in 0.5% (w/v) aqueous $H_2O_2$ and directional steering under a two-dimensional magnetic field. By integrating magnetic actuation with vision-based feedback, a custom-developed Python control program, and a custom-trained YOLOv11 object-detection model, we established the VIMAN framework for closed-loop autonomous microbot navigation.

The microbots were precisely guided along a range of prescribed complex trajectories and user-defined free-hand paths using waypoint-based closed-loop control, with more than 90% of the measured trajectory deviations remaining below 6 μm across the tested path geometries. The framework was subsequently extended to autonomous navigation in physically confined and dynamically changing environments. In a physically confined maze, the microbot autonomously followed a generated collision-free path toward the designated destination. In dynamic environments, real-time A* path replanning enabled the microbot to continuously adapt its navigation path while avoiding multiple actively swimming $Cu_2O/SiO_2$ Janus micromotors. The same framework was further extended to task-oriented cargo manipulation, enabling selective and sequential capture of multiple 2 μm PLA cargo particles, and their autonomous transport and delivery to a predefined location while continuously avoiding dynamic obstacles, with additional experiments demonstrating capture and transport of 2 μm silica ($SiO_2$) particles.

Overall, this work establishes an integrated vision-guided framework that combines programmable trajectory control, environment-aware path planning, real-time adaptive navigation, and task execution within a single light-driven and magnetically steerable Janus microbot platform. The developed approach provides a versatile basis for future autonomous microscale operations in lab-on-a-chip and organ-on-a-chip systems, microfluidic environments, and in vitro transport and delivery applications.

## Acknowledgements

The author D. P. Singh gratefully acknowledges funding support from Anusandhan National Research Foundation (ANRF) and the Ministry of Education-STARS program through the research grants ANRF/ARG/2025/003186/ENS and STARS-2/2023-0558, respectively. The author, K. C. Andia, is grateful to the DST-INSPIRE (No: DST/INSPIRE Fellowship/IF210642), government of India, for its financial support. The authors sincerely thank the Central Instrumentation Facility (CIF) at IIT Bhilai for their assistance with material characterization. K. C. Andia gratefully acknowledges Rajrupa Banerjee, Department of Physics, IIT Bhilai, for her valuable suggestions and insightful discussions on equation-based path planning.

## Data Availability Statement

Supporting Information is available upon reasonable request from the corresponding author.

## References

[1] X. Chen, C. Zhou, W. Wang, *Chem. – Asian J.* **2019**, *14*, 2388.
[2] S. Chen, D. E. Fan, P. Fischer, A. Ghosh, K. Göpfrich, R. Golestanian, H. Hess, X. Ma, B. J. Nelson, T. Patiño Padial, J. Tang, K. Villa, W. Wang, L. Zhang, A. Sen, S. Sánchez, *Nat. Nanotechnol.* **2025**, *20*, 990.
[3] C. Simó, M. Serra-Casablancas, A. C. Hortelao, V. Di Carlo, S. Guallar-Garrido, S. Plaza-García, R. M. Rabanal, P. Ramos-Cabrer, B. Yagüe, L. Aguado, L. Bardia, S. Tosi, V. Gómez-Vallejo, A. Martín, T. Patiño, E. Julián, J. Colombelli, J. Llop, S. Sánchez, *Nat. Nanotechnol.* **2024**, *19*, 554.
[4] A. Abbas, R. Mundaca-Uribe, L. Zhang, J. Wang, *Trends Biotechnol.* **2025**, *43*, 2197.
[5] K. C. Andia, H. Kumar, D. P. Singh, *Small* **2026**, e74901.
[6] C. K. Schmidt, M. Medina-Sánchez, R. J. Edmondson, O. G. Schmidt, *Nat. Commun.* **2020**, *11*, 5618.
[7] C. M. Maier, M. A. Huergo, S. Milosevic, C. Pernpeintner, M. Li, D. P. Singh, D. Walker, P. Fischer, J. Feldmann, T. Lohmüller, *Nano Lett.* **2018**, *18*, 7935.
[8] H. Ceylan, J. Giltinan, K. Kozielski, M. Sitti, *Lab. Chip* **2017**, *17*, 1705.
[9] A. Ghosh, P. Fischer, *Nano Lett.* **2009**, *9*, 2243.
[10] W. F. Paxton, K. C. Kistler, C. C. Olmeda, A. Sen, S. K. St. Angelo, Y. Cao, T. E. Mallouk, P. E. Lammert, V. H. Crespi, *J. Am. Chem. Soc.* **2004**, *126*, 13424.
[11] P. Sharan, Z. Xiao, V. Mancuso, W. E. Uspal, J. Simmchen, *ACS Nano* **2022**.
[12] J. Palacci, S. Sacanna, A. P. Steinberg, D. J. Pine, P. M. Chaikin, *Science* **2013**, *339*, 936.
[13] D. P. Singh, U. Choudhury, P. Fischer, A. G. Mark, *Adv. Mater.* **2017**, *29*, 1701328.

[14] V. R. A. Ferreira, M. A. Azenha, *Molecules* **2024**, *29*, 1154.
[15] S. K. Panda, S. Debata, K. C. Andia, S. Das, D. P. Singh, *Adv. Opt. Mater.* **2024**, *12*, 2400590.
[16] H. Zhou, C. C. Mayorga-Martinez, S. Pané, L. Zhang, M. Pumera, *Chem. Rev.* **2021**, *121*, 4999.
[17] U. Bozuyuk, O. Yasa, I. C. Yasa, H. Ceylan, S. Kizilel, M. Sitti, *ACS Nano* **2018**, *12*, 9617.
[18] M. Liu, Y. Wang, Y. Kuai, J. Cong, Y. Xu, H.-G. Piao, L. Pan, Y. Liu, *Small* **2019**, *15*, 1905446.
[19] Z. Wu, Y. Zhang, N. Ai, H. Chen, W. Ge, Q. Xu, *Adv. Intell. Syst.* **2022**, *4*, 2100266.
[20] A. Ghosh, D. Dasgupta, M. Pal, K. I. Morozov, A. M. Leshansky, A. Ghosh, *Adv. Funct. Mater.* **2018**, *28*, 1705687.
[21] L. Ren, N. Nama, J. M. McNeill, F. Soto, Z. Yan, W. Liu, W. Wang, J. Wang, T. E. Mallouk, *Sci. Adv.* **2019**, *5*, eaax3084.
[22] A. Aghakhani, O. Yasa, P. Wrede, M. Sitti, *Proc. Natl. Acad. Sci.* **2020**, *117*, 3469.
[23] M. Sokolich, Y. Yang, S. Cherukumilli, F. C. Kirmizitas, S. Das, *Adv. Robot. Res.* **2026**, e202500195.
[24] S. Gangwal, O. J. Cayre, M. Z. Bazant, O. D. Velev, *Phys. Rev. Lett.* **2008**, *100*, 058302.
[25] J. Yan, M. Han, J. Zhang, C. Xu, E. Luijten, S. Granick, *Nat. Mater.* **2016**, *15*, 1095.
[26] M. Ye, Y. Zhou, H. Zhao, Z. Wang, B. J. Nelson, X. Wang, *Adv. Intell. Syst.* **2023**, *5*, 2300311.
[27] R. K. Vardhan, M. Kumar, J. Xavier, *Nanophotonics* **2025**, *14*, 4821.
[28] F. Soto, E. Karshalev, F. Zhang, B. Esteban Fernandez De Avila, A. Nourhani, J. Wang, *Chem. Rev.* **2022**, *122*, 5365.
[29] J. Jiang, Z. Yang, A. Ferreira, L. Zhang, *Adv. Intell. Syst.* **2022**, *4*, 2100279.
[30] K. C. Andia, S. K. Panda, S. Mandal, S. Das, D. P. Singh, *Colloids Surf. Physicochem. Eng. Asp.* **2026**, *736*, 139606.
[31] D. P. Singh, W. E. Uspal, M. N. Popescu, L. G. Wilson, P. Fischer, *Adv. Funct. Mater.* **2018**, *28*, 1706660.
[32] Q. Cheng, X. Lu, Y. Tai, T. Luo, R. Yang, *ACS Biomater. Sci. Eng.* **2024**, *10*, 5562.
[33] S. Debata, S. Sahu, S. K. Panda, D. P. Singh, *J. Mater. Chem. B* **2024**, *12*, 2150.
[34] W. Yang, X. Wang, Z. Wang, W. Liang, Z. Ge, *Opt. Lasers Eng.* **2023**, *161*, 107380.
[35] J. G. Lee, R. R. Raj, N. B. Day, C. W. Shields, *ACS Nano* **2023**, *17*, 14196.
[36] S. Palagi, P. Fischer, *Nat. Rev. Mater.* **2018**, *3*, 113.
[37] É. O'Neel-Judy, D. Nicholls, J. Castañeda, J. G. Gibbs, *Small* **2018**, *14*, 1801860.
[38] S. Palagi, D. P. Singh, P. Fischer, *Adv. Opt. Mater.* **2019**, *7*, 1900370.
[39] L. Yang, J. Jiang, F. Ji, Y. Li, K.-L. Yung, A. Ferreira, L. Zhang, *Nat. Mach. Intell.* **2024**, *6*, 605.
[40] L. E. Beaver, M. Sokolich, S. Alsalehi, R. Weiss, S. Das, C. Belta, *IEEE Robot. Autom. Lett.* **2024**, *9*, 1819.
[41] J. Jiang, Z. Yang, A. Ferreira, L. Zhang, *Adv. Intell. Syst.* **2022**, *4*, 2100279.
[42] F. C. Kirmizitas, S. Debata, B. Wu, K. M. Ellwood, S. Cherukumilli, B. Ravenscraft, D. P. Rivas, M. Sokolich, A. Dutta, S. Das, *Adv. Healthc. Mater.* **2026**, *15*, e04779.
[43] F. M. Yurtsever, M. C. Ramakrishnan, I. K. Ross, I. Zelinka, H. Shahsavan, M. Otyepka, M. Pumera, *Adv. Mater.* **2026**, e74721.
[44] M. Sokolich, D. Rivas, Z. H. Shah, S. Das, *MRS Adv.* **2023**, *8*, 1005.
[45] T. Li, X. Chang, Z. Wu, J. Li, G. Shao, X. Deng, J. Qiu, B. Guo, G. Zhang, Q. He, L. Li, J. Wang, *ACS Nano* **2017**, *11*, 9268.

[46] F. C. Kirmizitas, D. P. Rivas, M. Sokolich, J. M. McNeill, A. Dutta, S. Das, *Adv. Intell. Syst.* **2025**, *7*, e202500300.
[47] H. M. Gauri, R. Patel, N. S. Lombardo, M. A. Bevan, B. Bharti, *Small* **2024**, *20*, 2403007.
[48] I. Rachbuch, S. Park, Y. Katz, T. Miloh, G. Yossifon, *Nat. Commun.* **2026**, *17*, 5993.
[49] X. An, S. Luo, H. Zhang, Q. Yang, Y. Ma, B. Wang, J. Du, Q. Wang, *Nat. Mach. Intell.* **2026**, *8*, 955.
[50] S. Palagi, D. P. Singh, P. Fischer, *Adv. Opt. Mater.* **2019**, *7*, DOI 10.1002/adom.201900370.
[51] K. Yuan, J. Bujalance-Fernández, B. Jurado-Sánchez, A. Escarpa, *Microchim. Acta* **2020**, *187*, 581.
[52] I. Buttinoni, G. Volpe, F. Kümmel, G. Volpe, C. Bechinger, *J. Phys. Condens. Matter* **2012**, *24*, 284129.
[53] K. Villa, F. Novotný, J. Zelenka, M. P. Browne, T. Ruml, M. Pumera, *ACS Nano* **2019**, *13*, 8135.
[54] P. Mayorga-Burrezo, C. C. Mayorga-Martinez, M. Pumera, *J. Colloid Interface Sci.* **2023**, *643*, 447.
[55] X. Fu, Z. Huang, G. Zhang, W. Wang, J. Wang, *PeerJ Comput. Sci.* **2025**, *11*, e2691.
[56] Q. Cheng, X. Lu, Y. Tai, T. Luo, R. Yang, *ACS Biomater. Sci. Eng.* **2024**, *10*, 5562.
[57] S. Hermanová, M. Pumera, *ACS Nanosci. Au* **2022**, *2*, 225.
[58] H. Kumar, S. K. Panda, K. C. Andia, Q. Ngo, W. Uspal, D. P. Singh, *Langmuir* **2026**, *42*, 13505.
[59] Z. Zhan, F. Wei, J. Zheng, W. Yang, J. Luo, L. Yao, *Nanotechnol. Rev.* **2018**, *7*, 555.
[60] D. P. Singh, U. Choudhury, P. Fischer, A. G. Mark, *Adv. Mater.* **2017**, *29*, 1701328.